\documentclass[letterpaper,11pt]{article}

\usepackage{amssymb}
\usepackage{amsmath}
\usepackage{amsthm}
\usepackage{stackrel}
\usepackage{graphicx}
\usepackage{authblk}
\usepackage{xcolor}
\usepackage{color}
\usepackage{float}
\usepackage{booktabs,siunitx}
\usepackage{multirow}
\usepackage{fullpage}
\usepackage{bm}
\usepackage{setspace}
\usepackage[utf8x]{inputenc}
\usepackage{algpseudocode, algorithm, algorithmicx}
\usepackage{romannum}
\usepackage{multirow}
\usepackage{soul}
\usepackage{comment}
\usepackage{cleveref}
\usepackage{url}

\usepackage{caption}
\usepackage{subcaption}

\DeclareMathOperator*\bigcircop{\bigcirc}

\newcommand\amsclass[1]{\hspace{9mm}
   \textbf{MSC codes. }#1}

\newcommand\C[1]\null

\title{Learning In-between Imagery Dynamics via Physical Latent Spaces}
\author{Jihun Han\thanks{jihun.han@dartmouth.edu}, Yoonsang Lee\thanks{yoonsang.lee@dartmouth.edu}, and Anne Gelb\thanks{Anne.E.Gelb@dartmouth.edu}}
\affil{Department of Mathematics, Dartmouth College}
\date{}

\begin{document}
\pagenumbering{arabic}

\maketitle
\begin{abstract}
We present a framework designed to learn the underlying dynamics between two images observed at consecutive time steps. The complex nature of image data and the lack of temporal information pose significant challenges in capturing the unique evolving patterns. Our proposed method focuses on estimating the intermediary stages of image evolution, allowing for interpretability through latent dynamics while preserving spatial correlations with the image. By incorporating a latent variable that follows a physical model expressed in partial differential equations (PDEs), our approach ensures the interpretability of the learned model and provides insight into corresponding image dynamics. We demonstrate the robustness and effectiveness of our learning framework through a series of numerical tests using geoscientific imagery data.
\end{abstract}
\amsclass{37M05, 62F99, 68T45}



\section{Introduction}
Understanding image dynamics from a set of complex measurement data is important in many applications, from the diagnosis or monitoring of a disease done by analyzing a series of medical (e.g.~MRI or ultrasound) images, \cite{xu2019deep},  to the interpretation of a sequence of satellite images used to study climate changes, natural disaster, or environmental conditions \cite{andersson2021seasonal}.
Here an ``image'' refers to a high-dimensional data frame that contains complex and condensed information within each pixel where these pixels are also spatially correlated. To understand the underlying dynamics between sequential images, therefore, it is essential to simultaneously decipher the intertwined relationship among their spatial and temporal features. 

A common approach for understanding such spatio-temporal dynamics involves the employment of physical models such as differential equations (DEs). By using the observed data to estimate the parameters in these corresponding DEs, it is possible to gain physical insights regarding their evolution \cite{guo2006identification, rudy2019data}.
However, directly applying such techniques to image dynamics is of limited use due to the intricate description that would be required by a suitable prior model, the highly nonlinear relationship among pixels, and the computational complexities arising from the high dimensionality of the images.
 
Deep learning methods explore the data representation that distills meaningful information, often referred to as features or latent variables, in order to efficiently and effectively address supervised tasks. Data embedding into the latent spaces (i.e., the spaces of latent variables) is acquired through the training of neural networks, the design of which is tailored to the specific data types or objectives. Methods employing convolutional neural networks (CNNs) have been shown to be capable of extracting informative spatial features from images, while recurrent neural networks (RNNs) such as LSTM (Long Short-Term Memory) \cite{hochreiter1997long} or GRU \cite{chung2014empirical} are widely used for identifying temporal patterns. One well-known method that combines these ideas is the Convolutional LSTM \cite{shi2015convolutional}. New techniques have improved upon this model by prescribing different combinations  \cite{wang2018predrnn++}. Some alternative approaches have also been introduced to address the uncertainty of future frames by including stochastic models \cite{babaeizadeh2017stochastic, denton2018stochastic} or generative adversarial network (GAN) models \cite{mathieu2015deep, saito2017temporal}. 

While these aformentioned methods incorporate standard feature extraction akin to those used in standard RNNs or GANs, it is possible to explore more organized feature or latent spaces that take into account the fundamental elements of image dynamics. This approach is commonly referred to as the disentanglement of visual representation.  It includes the factorization of each image into a stationary and temporally varying component \cite{denton2017unsupervised}, the elements for extrapolation that seamlessly propagate within frames and elements for generation that account for occlusion effects \cite{gao2019disentangling}, and the separation of motion and contents, each independently capturing the essential spatial layout of an image and salient objects, along with their corresponding temporal dynamics \cite{villegas2017decomposing}. A linear disentanglement with physical dynamics and residual components approach in which the learnable differential equation governs the physical dynamics was developed in \cite{guen2020disentangling}.

Although significant progress has been made in learning image dynamics when abundant temporal information is available for training, it is much more difficult to learn these same dynamics when the observable data are more limited. In practice, however, it can be hard to acquire a smooth time-lapse image sequence, particularly when dealing with satellite imagery of polar regions. These images often possess high spatial resolution but are temporally sparse due to the disparity in the time scale between sea ice deformation and the satellite's measurement cycle. Consequently, a compelling problem to address is the estimation of the intermediate dynamics between consecutive images, commonly known as temporal up-sampling, which serves to motivate the current investigation.
Some other applications include video interpolation for frame rate conversion, action recognition and motion tracking for surveillance and gesture recognition,  image morphing in visual effects, and various face manipulation applications \cite{zhang20023}.

In this study we present a novel approach to estimate the intermediate stage of scientific imagery. Our proposed method focuses on spatially-gridded measurement data, which heavily relies on various environmental state variables, including temperature, wind speed, and depth of snow, among others. We note that our framework is not limited to specific types of images and is generalizable to other applications. 
In particular we  propose a new machine learning framework to estimate the intermediate evolution stages of consecutive images. Motivated by the complex nature of images as measurements of environmental states, our goal is to uncover their hidden physical evolutions by transforming the observable images into relevant latent space variables. Despite limited temporal information from the data, we are able to utilize partial differential equation (PDE) models to learn latent dynamics that match both the initial and terminal states. Specifically, rather than directly attempting to learn image dynamics as in \cite{de2019deep}, we anatomize the data to obtain a simpler, more tractable, and explainable representations of the image dynamics. This approach is analogous to the method of characteristics for solving PDEs, which reduces them to simple ODE problems \cite{evans2022partial}.  It is also closely related to Koopman operator theory, which employs linear approximations of strongly nonlinear dynamic systems \cite{lusch2018deep}. The effectiveness of incorporating PDE models in machine learning has been validated in sequence-to-sequence problems \cite{gao2019disentangling, de2019deep}. In \cite{de2019deep}, the dynamics of SST (sea surface temperature) follow the advection-diffusion equation, and the advection vector fields for forward prediction are learned from historical sequences. In \cite{gao2019disentangling},  a recurrent model is designed to disentangle physical dynamics as driven by a learnable PDE model, thus effectively learning temporal dynamics for extrapolation. 

Our new approach, which we will refer to as  {\em latent space dynamics}, designs  the latent space to preserve spatial correlations between images and latent variables. The resulting latent dynamics are driven by the appropriately chosen PDEs and can be utilized to understand the original image dynamics. Our method efficiently scans spatial information from the images through continuously-sliding patches and feeds this information into the neural network components of our algorithms, effectively extracting spatial features and temporal information to drive the PDE models. Our model takes into account not only local observations but also global features through the common neural network architecture. Moreover, it takes advantage of adapting PDE models, which offer flexibility in incorporating prior knowledge of dynamics through effective regularization terms.

The remainder of the paper is organized as follows.  Section \ref{sec:problem_setup} provides some preliminary information and motivation regarding  the problem setup for in-between image dynamics. Section \ref{sec:main_algorithm} introduces the learning framework utilizing physical latent spaces, providing a detailed description of the training strategy. To validate the effectiveness of the proposed method, Section \ref{sec:numerical_results} presents numerical experiments conducted with geoscientific imagery data. Finally, in Section \ref{sec:conclusion} we conclude with discussion about the limitations of the current study and outline potential directions for future research.

\section{Learning in-between image dynamics}\label{sec:problem_setup}
Because it can provide insights into temporal dynamics and motion patterns of visual content, learning image dynamics from sequential imaging data is becoming more paramount in a variety of scientific disciplines.  For example, analyzing diverse and representative time-sequential data such as satellite measurements or synthetic aperture radar (SAR) imagery can reveal the complex dynamics of natural phenomena like weather patterns, ocean currents, or land use changes, and further allow us to make robust and reliable predictions that are adaptable to different scenarios. 

Collecting or measuring image sequences with sufficiently dense time steps is not always feasible, however, particularly in fields like geoscience where data acquisition is often constrained by satellite orbiting schedules or data transmission limitations. These limitations hinder our ability to understand the full temporal dynamics of the visual content. Furthermore, even with more frequently observed imaging sequences, it can still be difficult to interpret the captured movement within the data, making it all the more problematic to understand the underlying implications of the visual transformations. To address these issues, this investigation develops a technique designed to capture the temporal changes occurring between two images at different times while also prioritizing  interpretability. To achieve this goal, we harness the power of deep learning techniques enriched with physical models. This synergistic approach allows us to attain a deeper understanding of image dynamics and extract valuable insights.

\subsection{Problem setup}\label{subsec:setup}
The problem of interest is defined as one in which we seek to accurately predict the intermediary evolutionary stages between two given images. More specifically, let $\bm{X}_{t_0},\bm{X}_{t_1}  \in \mathbb{R}^{H\times W \times C}$ be two images measured at successive time $t_0$ and $t_1$, where $H, W$ and $C$ respectively denote the height, width, and the number of channels. For example, $\bm{X}_t$ represents a SAR image at a specific geographic location obtained from a moving satellite within a $14\sim 16$ day period, with the channel representing the number of measurement polarizations such as $HH$ or $HV$. Our objective is to learn the dynamical map of the image $\Phi_t: \mathbb{R}^{H\times W \times C} \rightarrow \mathbb{R}^{H\times W \times C}$, $t \in [0, t_1-t_0]$ that satisfies the initial and terminal conditions
\begin{equation}
\label{eq:initfinal}
\Phi_{0}(\bm{X}_{t_0})=\bm{X}_{t_0}, ~~~ \Phi_{t_1-t_0}(\bm{X}_{t_0})=\bm{X}_{t_1}.
\end{equation}
In essence, the map $\Phi_t$, $t\in (0, t_1-t_0)$ captures the in-between dynamics of $\bm{X}_t$ during an unobserved time period. Consequently, $\{\Phi_t(\bm{X}_{t_0}): t \in (0, t_1-t_0)\}$ represents the evolution of image $\bm{X}_t$ over that time period. We note that this problem differs from the standard frame prediction, where a model learns the map $\Phi:\mathbb{R}^{N \times H \times W \times C} \rightarrow \mathbb{R}^{H \times W \times C}$, using historical sequences of length $N$, with $N>1$, and sufficient temporal data. In our case, we assume limited temporal information is available, but we have ample spatial information in the given data.

It is crucial to acknowledge the intrinsic limitations of the problems associated with the uniqueness of the dynamics. The complexity of image dynamics may prevent a clear and  deterministic mapping of one frame to another as variations can arise due to factors such as object movement or inherent randomness in physical phenomena. Accordingly, we aim to propose a  learning framework with the capability to quantitatively interpret the resulting dynamics,  along with the flexibility to incorporate both prior knowledge as well constraints regarding the dynamics of interest.

\subsection{Previous efforts using optimal transport theory}\label{subsec:previous}
One approach utilized  for learning a dynamical map given a temporal sequence of image data involves the use of optimal transport (OT) theory, and is essentially  focused on finding the most efficient way to transport one probability distribution to another. This efficiency is quantified by the Wasserstein distance, a measure of the transport cost in terms of mass and distance. Formally, when dealing with 1D discrete probability distributions $\mu_1$ and $\mu_2$ of dimension $N$, the Wassertein distance is defined through the optimization problem
\begin{equation}
\label{eq:wasserstein}
W(\mu_1,\mu_2)  = \min \limits_{\pi \in \Pi(\mu_1,\mu_2)} \sum \limits_{i,j=1}^{N}c_{ij}\pi_{ij}.
\end{equation}
 Here, $c_{ij}$ is the ground cost from the location $i$ to $j$ (typically the squared Euclidean distance in practice) and $\Pi(\mu_1,\mu_2)$ is the set of all positive $N \times N$ matrices with marginal distributions (i.e., column and row sums) equal to $\mu_1$ and $\mu_2$, respectively. The formulation in \eqref{eq:wasserstein} is readily extended to multiple dimensions, and its direct calculation for 2D images has computational cost $\mathcal{O}(N^3)$, where $N$ represents the total number of pixels in the image. Recent developments, such as entropy-regularized formulations, have yielded more efficient numerical methods for approximating \eqref{eq:wasserstein} \cite{cuturi2013sinkhorn}, rendering it computationally tractable. These advancements have led to widespread applications in various fields, among those fluid dynamics \cite{agueh2015optimal, saumier2015optimal} and material sciences \cite{xia2023existence}. OT is particularly well-suited for scenarios where the total mass or pixel intensity remains conserved throughout the evolution, as it accounts for transformations within probability distributions.

Although OT provides a guiding principle to help characterize  smooth evolutionary dynamics in a variety of applications, it also presents certain limitations in specific scenarios. Notably, the standard OT framework formulated in Euclidean space is not well suited to capture rotational movement. Moreover, when the total mass undergoes changes due to the generation or disappearance of mass within the image, or when the acquired data are noisy, the normalization pre-processing step required for the method may introduce artifacts or inaccuracies into the resulting dynamics. Finally, we point out that OT primarily addresses the redistribution of mass or pixel density within the image framework,  and this may not adequately describe scenarios involving objects moving in or out of the domain through the image boundaries.

In contrast to the applications using OT, which are constrained by presuming fundamental principles regarding the governing image dynamics, our method adopts a more flexible approach. Specifically, we impose a marginal assumption regarding the evolution of  the {\em image features}, as opposed to the images themselves. Moreover, rather than pre-define these features, we enable our framework to autonomously extract and explore various features and their corresponding dynamics, utilizing the image statistics as part of our machine learning approach. This results in diverse dynamic patterns in image representation, and is closely aligned with real-world scenarios. Such flexibility enables us to overcome the limitations that are potentially imposed by a pre-defined governing principle.

\section{Learning through physical latent spaces}\label{sec:main_algorithm}
Unlike continuous physical models that are regularly used to describe evolutionary dynamics, images capture a snapshot of multiple physical quantities simultaneously and represent them with pixel values. In other words, an image can be interpreted as a collection of spatially-gridded measurement data that incorporates several environmental state variables, such as temperature and wind speed, and in the case of SAR images on arctic regions, the depth of snow. Deciphering how each of the underlying physical quantities evolve and interact solely based on pixel value image data is non-trivial, as it requires a deep understanding of how variables within the image are related and represented. Inspired by this intuitive understanding between sequential image data and the underpinning physical quantities that are present in each image, we propose a learning framework aimed to uncover the evolution of images by considering the dynamics of the underlying variables, i.e.,~the relevant latent space variables, that follow established physical models. In so doing, we can better interpret how images change over time.

\subsection{Model flow framework}
We design the model for the objective map $\Phi_t$ with three components as  
\begin{subequations}
\label{eq:objectivemap}
\begin{align}
\Phi_t ~:~  \mathcal{I}& ~\rightarrow~  (\mathcal{P} \times \mathcal{Q}) ~\rightarrow~  \mathcal{P} ~\rightarrow~ \mathcal{I}, ~~\textrm{as} ~~ \Phi_t = \psi \circ  P_t \circ (\phi, \eta), ~t \in [0, t_1-t_0], 
\label{eq:phi_t}\\
&\mathcal{I} : \text{the image space for }  \bm{X}_t, t\in[t_0, t_1], 
\label{eq:spacesI}\\
&\mathcal{P} : \text{the latent space for } \bm{Z}_t, t \in[t_0, t_1], 
\label{eq:spacesP}\\
&\mathcal{Q} : \text{the space for physical model parameters } \bm{W}_t, t \in[0, t_1-t_0]. 
\label{eq:spacesQ}
\end{align}
\end{subequations}
Here $\phi$ and $\psi$ are the identification maps, also respectively referred to as the encoding and decoding maps between the image and latent space. The map $P_t$ represents the evolution of $\bm{Z}_t$ in the latent space and is driven by a predefined physical model as $\bm{Z}_{t_0+t}=P_t(\bm{Z}_{t_0})$ or $\bm{Z}_{t_0+t}=P_t(\bm{Z}_{t_0},\{\bm{W}_s\}_{0\leq s\leq t_1-t_0})$ along with the physical parameters $\bm{W}_t$ from the map $\eta$. The map $\phi$ plays a role in extracting the features $\bm{Z}_{t_0}$ from the initial image $\bm{X}_{t_0}$, which evolve to the latent state $\bm{Z}_{t_1}$ corresponding to the destination image $\bm{X}_{t_1}$. The intermediary stages of image $\bm{X}_t$, $t \in (0, t_1-t_0)$ are generated by pulling back from the corresponding latent variable $\bm{Z}_t$ through the decoding map $\psi$ as $\bm{X}_t = \psi(\bm{Z}_t)$. 

The main objective of the model is to search for an appropriate transformation from the image space to a feature space, such that the original problem in the image space can be effectively addressed within the framework of the predefined physical model. To achieve this, we learn the transformation by employing neural networks $\phi=\phi(\cdot; \bm{\theta}_1)$ and $\psi=\psi(\cdot; \bm{\theta}_2)$ along with $\eta=\eta(\cdot;\bm{\theta}_3)$ under the supervision with given image data $\bm{X}_{t_0}$ and $\bm{X}_{t_1}$ as
\begin{equation}
\label{eq:obj_map2}
\Phi_{t_1-t_0}(\bm{X}_{t_0};\{\bm{\theta}_1,\bm{\theta}_2, \bm{\theta}_3\}) = \psi(P_{t_1-t_0}(\phi(\bm{X}_{t_0};\bm{\theta}_1), \eta(\bm{X}_{t_0};\bm{\theta}_3));\bm{\theta}_2) = \bm{X}_{t_1}.
\end{equation}

To interpret the resulting dynamics of the image through latent correspondence, we make sure to design the transformation $\phi$ so that, at least to some extent, the spatial correlation between the image and latent variables is preserved. This can be achieved by designing the latent space $\mathcal{P}$ with the same or comparable spatial dimensionality as in the original image.  Further, the neural network is constructed using only convolution operators, that is, without including dense layers that are used in fully convolutional neural networks. Details regarding the network architectures and training are discussed in \Cref{sec:training_model}. 

It is important to point out that our approach differs from one that incorporates PDEs in sequence-to-sequence models.  Specifically, we avoid having to model constraints that arise when attempting to directly model a PDE in the image space \cite{de2019deep}, which can become particularly arduous when considering intricate measurements for various physical attributes. In particular, autonomously learning the form of PDEs on abstract lower-dimensional latent spaces \cite{guen2020disentangling} may introduce further limitations as the learned PDE may be too complex to interpret. Also, the relationship between the latent and image dynamics are unclear.

\subsection{Physical model in latent spaces} \label{sec:latent}
We utilize partial differential equations (PDEs) to model latent dynamics. In particular, we assume that the dynamics in the latent space can effectively explain changes in image space, and that we have the flexibility to select an appropriate PDE model that aligns with our specific  interest in image dynamics. For example, diffusion equations accurately represent temperature changes in a system, while advection equations capture wind patterns. Occasionally, a combination of advection and diffusion equations may be necessary to encompass the intricacies of the phenomenon.

In this work we focus on the advection equation as the primary physical model and aim to better understand the dynamics associated with geoscientific images. The advection vector fields provide detailed descriptions of the flow or motion of  physical variables. Consequently these fields offer valuable information about sea ice movement as well as its contributions to possible crack formation. By additionally incorporating information regarding the advection of buoys placed within the region of the SAR imagery, we can also increase our understanding of the ocean currents. 

\begin{figure}[!h]
\centering
\includegraphics[width=0.95\textwidth]{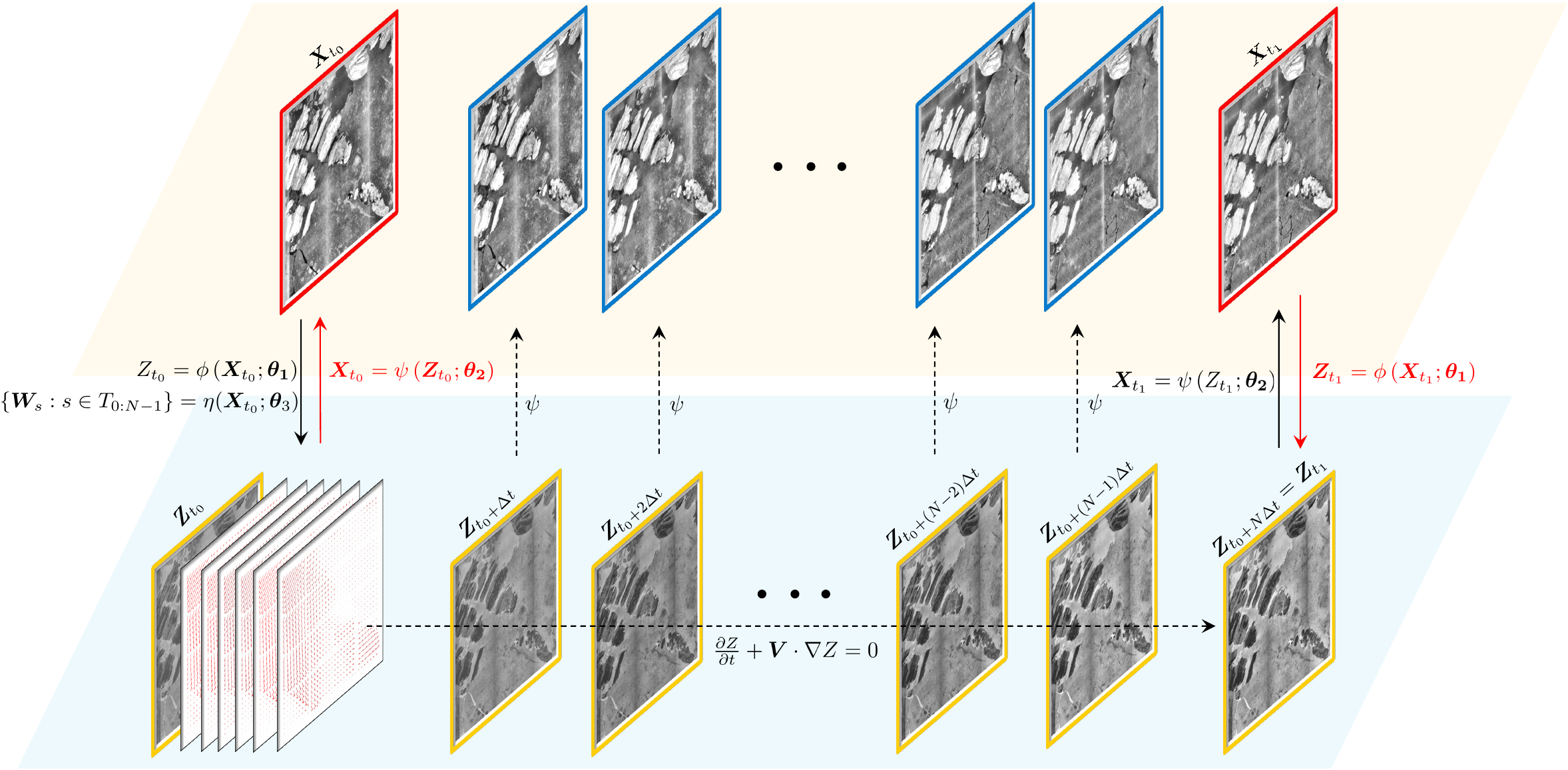}
\caption{Schematic diagram of the model to learn the dynamics of in-between images $\bm{X}_{t_0}$ and $\bm{X}_{t_1}$ (red borders). The model learns the latent space (through encoding $\phi$ and decoding $\psi$) where the latent variable $\bm{Z}_{t}$ (yellow borders) physically evolves during the time period $[t_0, t_1]$ and reaches to the state $\bm{Z}_{t_1}$ corresponding to the image $\bm{X}_{t_1}$. The intermediary images (blue borders) are recovered by decoding latent correspondence.}
\label{fig:model_flow}
\end{figure}

For ease of presentation, we consider the latent space $\mathcal{P} $ with the same spatial dimensionality as the image space $\mathcal{I}$, i.e. $\bm{Z}_t \in \mathcal{P} \subset \mathbb{R}^{H\times W \times 1}$ in \eqref{eq:objectivemap}, where $H$ and $W$ are the height and width dimensions (first described in \Cref{subsec:setup}). The temporal discretization of $[t_0,t_1]$ is given by \vspace{-1.8mm}
\begin{equation} 
T_{0:N} = \{t_0+j\Delta t\}_{j = 0}^N \nonumber
\end{equation}
where $\Delta t = \frac{t_1-t_0}{N}$, while the map $\eta$,  as defined in \eqref{eq:phi_t}, provides the $N$ vector fields $\bm{W}_t \in \mathcal{Q} \subset \mathbb{R}^{H\times W \times 2}$  to drive the advection dynamics of $\bm{Z}_t$ over the time domain. For numerical computation we consider the advection equation on a rectangular domain $\Omega \subset \mathbb{R}^2$ given by  \vspace{-1.8mm}
\begin{equation}
z: \Omega \times [t_0, t_1] \rightarrow \mathbb{R}, ~~~~ \frac{\partial z}{\partial t} + \bm{w}(\bm{x},t)\cdot \nabla z = 0,
\end{equation}
where $\bm{w}:\Omega \times [t_0, t_1] \rightarrow \mathbb{R}^2$ are the  advection vector fields. By conservation, the evolution of $z$ over the small time period $\Delta t$ is approximated by
\begin{equation}
z(\bm{x}, t+\Delta t) = z(\bm{x}-\bm{w}(\bm{x}, t)\Delta t, t).
\end{equation}
The evolution of $\bm{Z}_t$ driven by $\bm{W}_t$ is considered to be the spatially-discretized dynamics of the continuous quantity $z$ with respect to $\bm{w}$. To evaluate the evolution, we use interpolation on a discretized domain which can be described as follows:\vspace{-2mm}
\begin{itemize}
\setlength\itemsep{0.08em}
\item Define $\Omega_D$ to be the uniform  discretization of $\Omega$ with the same dimension $H\times W$ as the latent variable $\bm{Z}_t$ and $\Omega_D(\bm{w})=\{\bm{x}-\bm{w}\Delta t: \bm{x} \in \Omega_{D}\}$ to be the shifted grid points by the vector field $\bm{w}$. 
\item Compute $\bm{Z}_{t+\Delta t}$ by interpolating $\bm{Z}_t$ on $\Omega_D(\bm{w}(x,t))=\Omega(\bm{W}_t)$ at the uniform grid $\Omega_{D}$, which we simply write as \vspace{-2mm} 
\begin{equation*}
\bm{Z}_{t+\Delta t} = P_{\Delta t}(\bm{Z}_t; \bm{W}_t) =  \texttt{Interpolation}\left(\Omega_D; \left(\bm{Z}_t, \Omega_D(\bm{W}_t)\right)\right).
\end{equation*}
\item The overall advection of $\bm{Z}_t$ over the time period $[t_0, t_1]$ then consists of multiple stepping given by \vspace{-2mm}
\begin{equation*}
\bm{Z}_{t_1}=P_{t_1-t_0}(\bm{Z}_{t_0};\{\bm{W}_s: s\in T_{0:N-1} \}) = \left(\bigcircop_{j=0}^{N-1} P_{\Delta t}(\cdot; \bm{W}_{t+j\Delta t})\right)(\bm{Z}_{t_0}).
\end{equation*}
\end{itemize}
\Cref{table:summary of mappings} and \Cref{fig:model_flow} summarize the evolutions of the original model $\Phi_t=\psi \circ P_t \circ (\phi, \eta)$ along with the corresponding latent space advection model.

\begin{table}
\centering
\scalebox{0.95}{
\begin{tabular}{c | c | c} 
 & $\text{dimensionality}$ & $\text{maps}$   \\
\hline
\hline \rule{0pt}{3ex}  
$\text{encoding}$           & $\phi : \mathbb{R}^{H\times W \times C} \rightarrow \mathbb{R}^{H \times W \times 1}  $&  $\phi(\bm{X}_{t_0}; \bm{\theta}_1)=\bm{Z}_{t_0}$  \\
$\text{field generation}$   & $\eta : \mathbb{R}^{H\times W \times C} \rightarrow \mathbb{R}^{N \times H \times W \times 2}$  & $\eta(\bm{X}_{t_0}; \bm{\theta}_3)=\{\bm{W}_s : s\in T_{0:N-1}\}$ \\
$\text{advection}$        & $P_t : \mathbb{R}^{H\times W \times 1} \rightarrow \mathbb{R}^{H\times W \times 1}$ & $P_{t_1-t_0}(\bm{Z}_{t_0}; \{\bm{W}_s : T_{0:N-1}\}) = \bm{Z}_{t_1}$ \\
$\text{decoding}$         & $\psi : \mathbb{R}^{H\times W \times 1} \rightarrow \mathbb{R}^{H \times W \times C} $ & $\psi(\bm{Z}_{t_1}; \bm{\theta}_2)=\bm{X}_{t_1}$
\end{tabular}}
\caption{The summary of ingradients in the model $\Phi_t = \psi \circ  P_t \circ (\phi, \eta)$ }
\label{table:summary of mappings}
\end{table}

\subsection{Training the model}\label{sec:training_model}

\subsubsection{Learning patch-to-patch dynamics}
Simultaneously learning the diverse features and evolution of an image can be computationally challenging and inefficient, especially for large images. To address this issue, we allow the neural networks to locally learn the image through smaller windows (patches), which are then combined to obtain the overall image dynamics. Specifically, we consider patches of size $H_p \times W_p$, $H_p < H$, $W_p < W$, and continuously scan the entire images $\bm{X}_{t_0}$ and $\bm{X}_{t_1}$ to generate a training dataset consisting of initial and final patches $\left\{\left(\bm{X}^{(i)}_{t_0}, \bm{X}^{(i)}_{t_1}\right): \bm{X}_{t_0}^{(i)}\in \mathbb{R}^{H_p\times W_p\times C}, \bm{X}_{t_1}^{(i)}\in \mathbb{R}^{H_p\times W_p\times C}\right\}$, where $i = 1,\dots,N$ denotes the index of patch.  The objective of our model is to learn the dynamics between patches across all training data, and then combine these patch-to-patch dynamics to capture the overall dynamics of the entire image. It is important to note that while each patch focuses on a local region, the neural network implicitly considers global features of the entire image due to the continuous scanning of the patches, thus ensuring that information from various regions in the image is incorporated into the learning process.

\subsubsection{Neural networks}\label{subsubsec:neural_networks}

\begin{figure}[!h]
\centering
\includegraphics[width=0.95\textwidth]{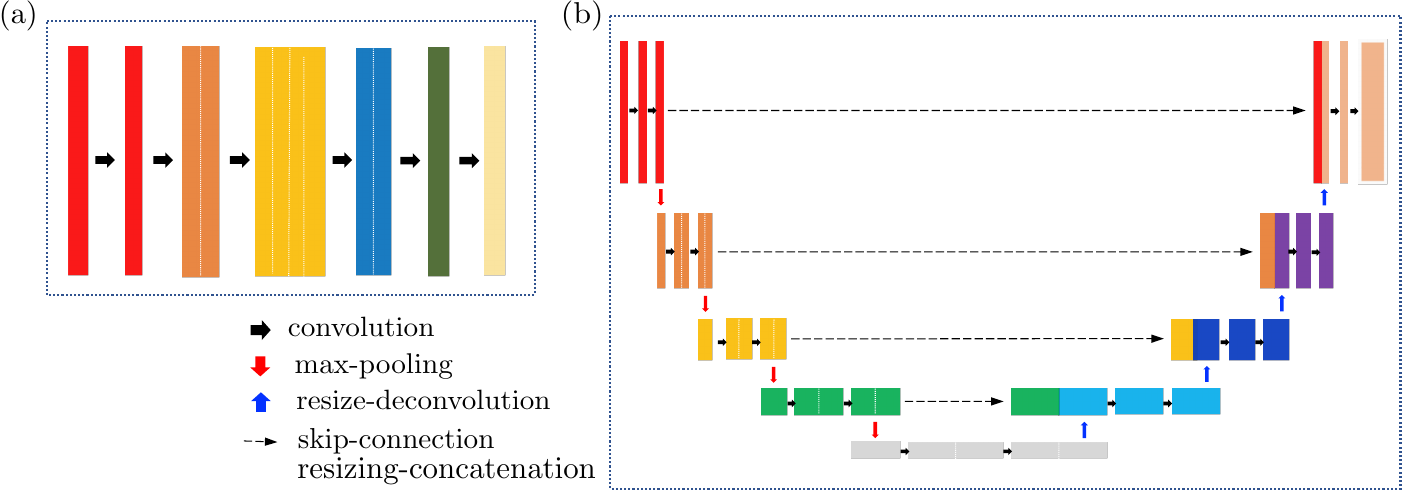}
\caption{Architectures of neural network ingredients in the proposed model. (a) the encoding map $\phi(\cdot;\bm{\theta}_1)$; stacks of only convolutional layers (b) the decoding map $\psi(\cdot;\bm{\theta}_2)$ and the advection field extraction map $\eta(\cdot;\bm{\theta}_3)$; U-net structure particularly with resize-deconvolutions and skip-connections by resizing-concatenation.}
\label{fig:networks}
\end{figure}

We learn the physical latent space (the encoding $\phi$ and the decoding $\psi$) and the advection field (the map $\eta$) through neural networks, with the aim to design the architecture for the encoding map $\phi(\cdot;\bm{\theta}_1)$ to preserve the spatial correlation between the image patch and corresponding latent variable. To achieve this goal, we stack the convolution layers as illustrated in \Cref{fig:networks}(a). Each pixel in the latent variable is determined solely by the local receptive field, i.e.,~the neighborhood around the corresponding location in the image, which is defined by the kernel size of convolution.\footnote{Here we use $3\times 3$ or $5\times 5$ for the neighborhood size, which will be discussed further in \Cref{sec:numerical_results}.}  Note that this approach contrasts with the inclusion of pooling layers or dense layers, which incorporate information from the entire image and result in non-transitive local correlations between the image and the variable, albeit with more condensed information from the overall image. For the decoding map $\psi(\cdot;\bm{\theta}_2)$ and advection field extraction $\eta(\cdot;\bm{\theta}_3)$, we employ the U-net \cite{ronneberger2015u} architecture shown in \Cref{fig:networks}(b). This U-net structure comprises a downsampling stage (also known as a feature extractor or encoder) and an upsampling stage (also known as a decoder), which are interconnected by skip connections that transfer information from the downsampled features to the upsampled correspondence. The original U-net, initially designed for biomedical image segmentation, employs deconvolutional upsampling followed by concatenation with the corresponding cropped features from the downsampling. In our experiments, however, we observed the presence of chessboard artifacts in the decoded images, as also documented in \cite{odena2016deconvolution}, as well as a contaminated advection field. To mitigate the presence of these artifacts during the upsampling process, we replace the deconvolution operators with resize-convolution \cite{odena2016deconvolution}. Additionally, to prevent the loss of information from the downsampled features, we replace the cropping operation with resizing, which allows us to concatenate the upsampled features while preserving the available information.

\subsubsection{Loss functions}\label{subsec:losses}
We design the loss functions for the model to learn the evolution from $\bm{X}_{t_0}$ to $\bm{X}_{t_1}$ while also providing the intermediary stages. To this end we first construct the loss, $\mathcal{L}_{\text{dynamics}}$,  to measure the discrepancy of the model output at final time $t_1$ from the given destination image $\bm{X}_{t_1}$ as
\begin{equation}\label{eq:dynamicsloss}
\mathcal{L}_{\text{dynamics}}(\bm{\theta}_1,\bm{\theta}_2, \bm{\theta}_3) = \sum \limits_{i} \left| \psi\left(P_{t_1-t_0}\left(\phi\left(\bm{X}^{(i)}_{t_0};\bm{\theta}_1\right) ; \eta\left(\bm{X}^{(i)}_{t_0};\bm{\theta}_3\right)\right);\bm{\theta}_2\right) - \bm{X}^{(i)}_{t_1} \right|^2,
\end{equation}
where the latent states at the initial and terminal times correspond respectively to
\[\bm{Z}^{(i)}_{t_0}=\phi\left(\bm{X}^{(i)}_{t_0};\bm{\theta}_1\right), \quad\quad
\bm{Z}^{(i)}_{t_1}=P_{t_1-t_0}\left(\bm{Z}^{(i)}_{t_0} ; \eta\left(\bm{X}^{(i)}_{t_0};\bm{\theta}_3\right)\right). \]

We note that \eqref{eq:dynamicsloss} is not sufficient to drive the latent dynamics of $\bm{Z}_t$ through a non-trivial vector field as it focuses on the reconstruction of the destination image $\bm{X}_{t_1}$ from the latent variable $\bm{Z}_{t_1}$.  To distinguish the latent variables $\bm{Z}_{t_0}$ and $\bm{Z}_{t_1}$ corresponding to the different images $\bm{X}_{t_0}$ and $\bm{X}_{t_1}$, the maps $\phi(\cdot;\bm{\theta}_1)$ and $\psi(\cdot;\bm{\theta}_2)$ should play role in the auto-encoder of the image space. This is possible under the reasonable assumption that the two images share the same latent space.  We define the loss function for the auto-encoder as
\begin{equation}
\label{eq:autoloss}
\mathcal{L}_{\text{AE}}(\bm{\theta}_1, \bm{\theta}_2) = \sum \limits_{i} \left|\bm{X}_{t_0}^{(i)} - \psi\left(\bm{Z}_{t_0}^{(i)};\bm{\theta}_2\right) \right|^2 + \left|\bm{Z}_{t_1}-\phi\left(\bm{X}^{(i)}_{t_1};\bm{\theta}_1 \right) \right|^2.
\end{equation}
We can also impose the conditions on the advection fields $\bm{W}_t$ for stabilizing the learning process or incorporating prior knowledge through the regularization loss term. Here we use $\ell_2$-regularization, which both stabilizes the learning process as well as smooths the spatial and temporal results, respectively as 
\[\mathcal{L}_{\text{magnitude}}(\bm{\theta}_3) = \sum \limits_{i}  \sum \limits_{t \in T_{0:N}}\left|\bm{W}^{(i)}_{t} \right|^2 , \quad\quad
\mathcal{L}_{\text{smooth}}(\bm{\theta}_3) = \sum \limits_{i}  \sum \limits_{t \in T_{1:N}}\left|\bm{W}^{(i)}_{t} - \bm{W}^{(i)}_{t-\Delta t} \right|^2.\]
In sum, we train our model using a weighted combination of loss functions given by
\begin{equation}
\label{eq:sumloss}
\mathcal{L}(\bm{\theta}_1,\bm{\theta}_2,\bm{\theta}_3) = \mathcal{L}_{\text{dynamics}} + \lambda_{\text{AE}}\mathcal{L}_{\text{AE}}+\lambda_{\text{magnitude}}\mathcal{L}_{\text{magnitude}} + \lambda_{\text{smooth}}\mathcal{L}_{\text{smooth}},
\end{equation}
which is optimized by a gradient descent method 
\begin{equation}
\bm{\theta}^{(n)} = \bm{\theta}^{(n-1)} - \alpha \nabla_{\bm{\theta}} \mathcal{L}\left(\bm{\theta}^{(n-1)}\right),~\bm{\theta} = \{\bm{\theta}_1,\bm{\theta}_2,\bm{\theta}_3\},~~ \alpha > 0,
\end{equation}
where $\alpha$ is a learning rate.

\section{Numerical results}\label{sec:numerical_results}
We now validate the robustness and efficacy of our new latent space dynamics learning framework that is designed to capture intricate inter-imagery dynamics from an initial and destination image using the model framework described by \eqref{eq:objectivemap}. Here we consider a suite of polar region SAR images from which we are specifically interested in investigating sea ice behavior. The environments we analyze involve macro-scaled sea ice deformations, and encompass combinations of translation, rotation, and fracture. Our primary goal is to quantify these deformations using vector fields derived from a physical model, while also offering qualitative insights regarding the underlying scenes.

For our data collection we access SAR image data from the Copernicus Open Access Hub. Each of these images includes multiple polarization values and is georeferenced in longitude-latitude coordinates. As a preprocessing step we convert the $HH$-polarized data into a rectangular coordinate system using projection techniques. We then rescale the data, ensuring that each pixel corresponds to a substantial geographical distance of $0.5 \sim 1$ kilometers.

In all of our experiments we fix the spatial dimension of the patch as dimensions $256 \times 256$. The image is systematically scanned using consecutive overlapping patches with stride parameters denoted by 
  $S_H$ (height) and $S_W$ (width),\footnote{For example, the scanning is achieved with strides $S_H$ and $S_W$ means that the centers of two consecutive patches are located $S_H$ vertical pixels and $S_W$ horizontal pixels apart.} which we will specify later, to constitute our training dataset. To help facilitate input to the neural networks, we normalize each patch to a range of $[-1, 1]$. When exploring latent patch-to-patch dynamics, we define the latent domain $\Omega$ as $[0, 1]^2$, and align its discretization to be consistent with the patch dimensions ($256 \times 256$). Finally, we introduce $N_\text{evolution}$ stages for the dynamics, each evolving over a time step $\Delta t = 0.1$.

Our neural network architecture, as detailed in Section \ref{subsubsec:neural_networks}, is structured as follows:\vspace{-2mm}
\begin{itemize}
\setlength\itemsep{0.1em}
\item  The encoding map $(\phi)$ comprises six hidden convolutional layers, each using $3\times 3$ kernel and applying a leaky ReLU activation function with a parameter of $0.2$.
\item Decoding $(\psi)$ and field extraction $(\eta)$ maps adopt U-net structures with three downsampling and upsampling stages. Each stage incorporates two hidden convolutional layers, followed by a leaky ReLU activation function with a parameter of 0.2. We use $3\times 3$ convolutional kernels for decoding and $5\times 5$ for field extraction. Max-pooling is applied during each downsampling step.
\end{itemize}\vspace{-2mm}

We use the stochastic gradient descent (SGD) method along with the Adam optimizer to perform optimization for the neural networks \cite{kingma2014adam}. The learning parameters $\beta_1$ and $\beta_2$ are set to $0.9$ and $0.999$, respectively, for efficient convergence. We adopt an exponential decay learning rate, initially defined as $\alpha$, with a decay rate of $\gamma$ applied every $10000$ training iterations. The specific hyperparameters pertaining to our learning framework will be detailed for each individual test problem.

Finally, to ensure a comprehensive evaluation of our framework's performance, we compare the dynamics generated by our approach with the results obtained through two alternative methods: (1) an optimal transport approach \cite{zhou2022efficient};  and (2) a direct application of the PDE model in the image space without passing through latent spaces (i.e., $\mathcal{P} = \mathcal{I}$ with the identity encoding and decoding $\phi(\bm{X})=\bm{X}$ and $\psi(\bm{Z})=\bm{Z}$ in our method).\footnote{Animated visualization of the estimated dynamics can also be found at \\ \mbox{\url{https://sites.google.com/view/jihun-han/home/research-gallery/learning-dynamical-systems}.}}

\subsection{Transition dynamics}\label{subsec:transition}

\begin{figure}[h!]
\centering
\includegraphics[width=1.0\textwidth]{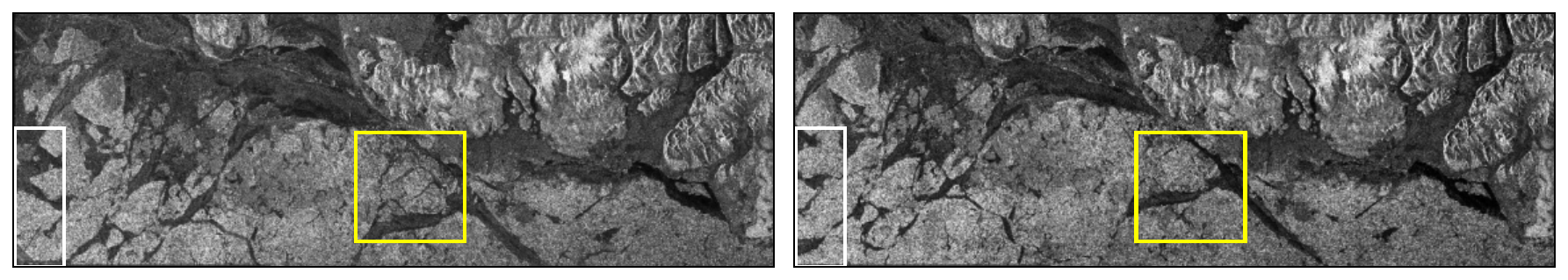}
\caption{Initial (left) and terminal (right) images for Example 1 captured at a 10-day interval. The colored bounding boxes indicate the regions of interest.}
\label{fig:drift_image_initial_and_terminal}
\end{figure}

\begin{table}[h!]
\centering
\scalebox{0.95}{
\begin{tabular}{c || c | c | c} 
 & $\text{encoder}$ $\phi(\cdot;\bm{\theta}_1)$ & $\text{decoder}$ $\psi(\cdot;\bm{\theta}_2)$ & $\text{field extraction}$ $\eta(\cdot;\bm{\theta}_3)$   \\
\hline
\hline \rule{0pt}{3ex}  
$\text{hidden layers}$& $32, 64, 128, 64, 32, 16, 8$  & $\cdot$ & $\cdot$   \\
\hline
$\text{input layers}$ & $\cdot$ & $16$ &  $16$  \\
$\text{downsampling}$ & $\cdot$ & $16(2), 32(2), 64(2)$ & $16(2), 32(2), 64(2)$ \\
$\text{bottleneck}$   & $\cdot$ & $128$ & $128$ \\
$\text{upsampling}$   & $\cdot$ & $64(2), 32(2), 16(2)$ & $64(2), 32(2), 16(2)$ \\
$\text{output layers}$& $\cdot$ & $16$  & $16$  \\
\end{tabular}}
\newline
\vspace*{0.3cm}

\scalebox{0.95}{
\begin{tabular}{c | c | c | c | c | c | c} 
$N_{\text{evolution}}$ & $S_H \times S_W$ & $\lambda_{\text{AE}}$ & $\lambda_{\text{magnitude}}$ & $\lambda_{\text{smooth}}$ & $\alpha$ & $\gamma$ \\
\hline
\hline 
$10$ & $1 \times 1$ & $1.$ & $0.01$ & $0.01$ &  $0.0001$ & $0.8$ \\
\end{tabular}}
\caption{Hyperparameters for training the proposed model in Example 1. (Top) the numbers of convolutional filters in neural networks; (Bottom) number of evolution steps, stride for scanning the patch over the whole image, regularization parameters, and learning parameters.}
\label{table:hyperparameters for example 1}
\end{table}

As our first test case, we apply our new method to learn transition dynamics (subsequently referred to as {\bf Example 1}). \Cref{fig:drift_image_initial_and_terminal} shows image data captured at a $10$ day interval. These images reveal the progression of sea ice conditions, and in particular we observe changes along the shoreline in the lower right quadrant adjacent to the unchanging land visible in the upper left quadrant.

We model the intermediate stages of evolution between these images by approximating change on a daily basis, that is using $N_{\text{evolution}}=10$. We scan the images measuring $256\times 1024$ pixels using  patches with stride $1\times 1$, resulting in $768$ pairs of overlapping patches, which are then used to train our model with the hyperparameters provided in Table~\ref{table:hyperparameters for example 1}. The trained model is subsequently employed to infer dynamics from patch to patch at the intermediate stages for prescribed non-overlapping areas, yielding comprehensive coverage of the entire images.\footnote{Indeed, the desired dynamics information can be prescribed at pixel locations using any number of patch to patch combinations.  Here for simplicity we use the non-overlapping patches to cover the domain.} 

\begin{figure}[H]
\centering
\includegraphics[width=0.87\textwidth]{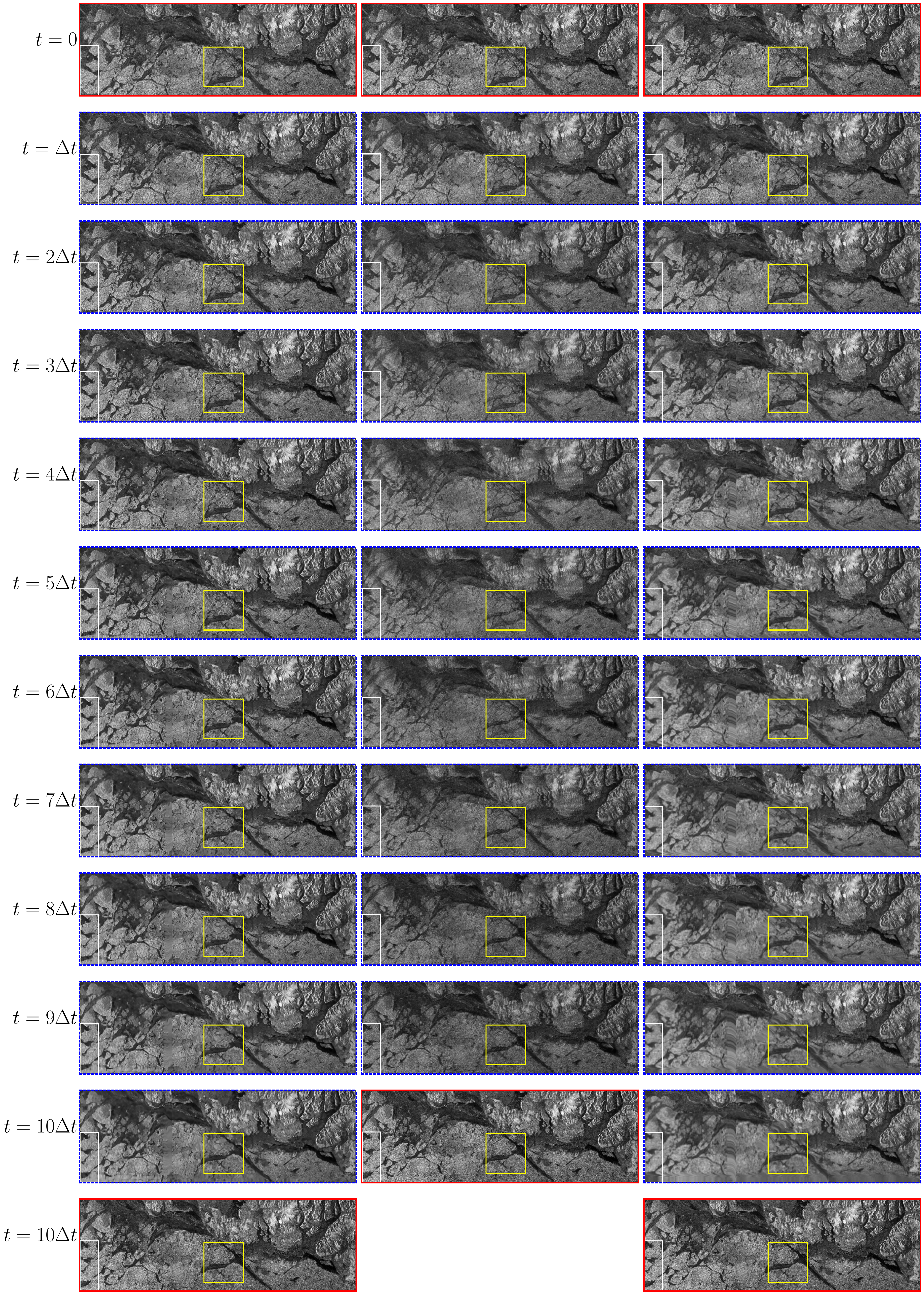}
\caption{The estimation for intermediate stages (blue dashed borders) of transition-dominant dynamics given initial and terminal states (red borders) for Example 1.  Each row marks a different time $t = j\Delta t$, $j = 0,\dots,10$. (Left)  our new latent space dynamics approach; (middle)  OT \cite{zhou2022efficient}; and (right) direct application of PDE model in the image space.}
\label{fig:drift_all_methods}
\end{figure}

\begin{figure}[H]
\begin{subfigure}[b]{1.0\textwidth}
\centering
\includegraphics[width=1\textwidth]{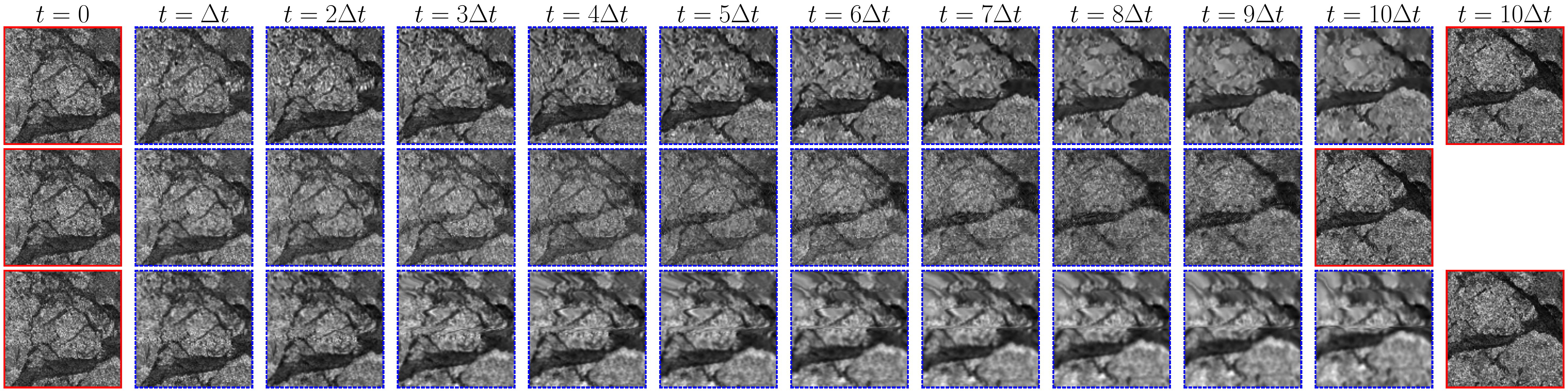}
\caption{The region enclosed by yellow bounding box}
\end{subfigure}
\begin{subfigure}[b]{1.0\textwidth}
\centering
\includegraphics[width=1\textwidth]{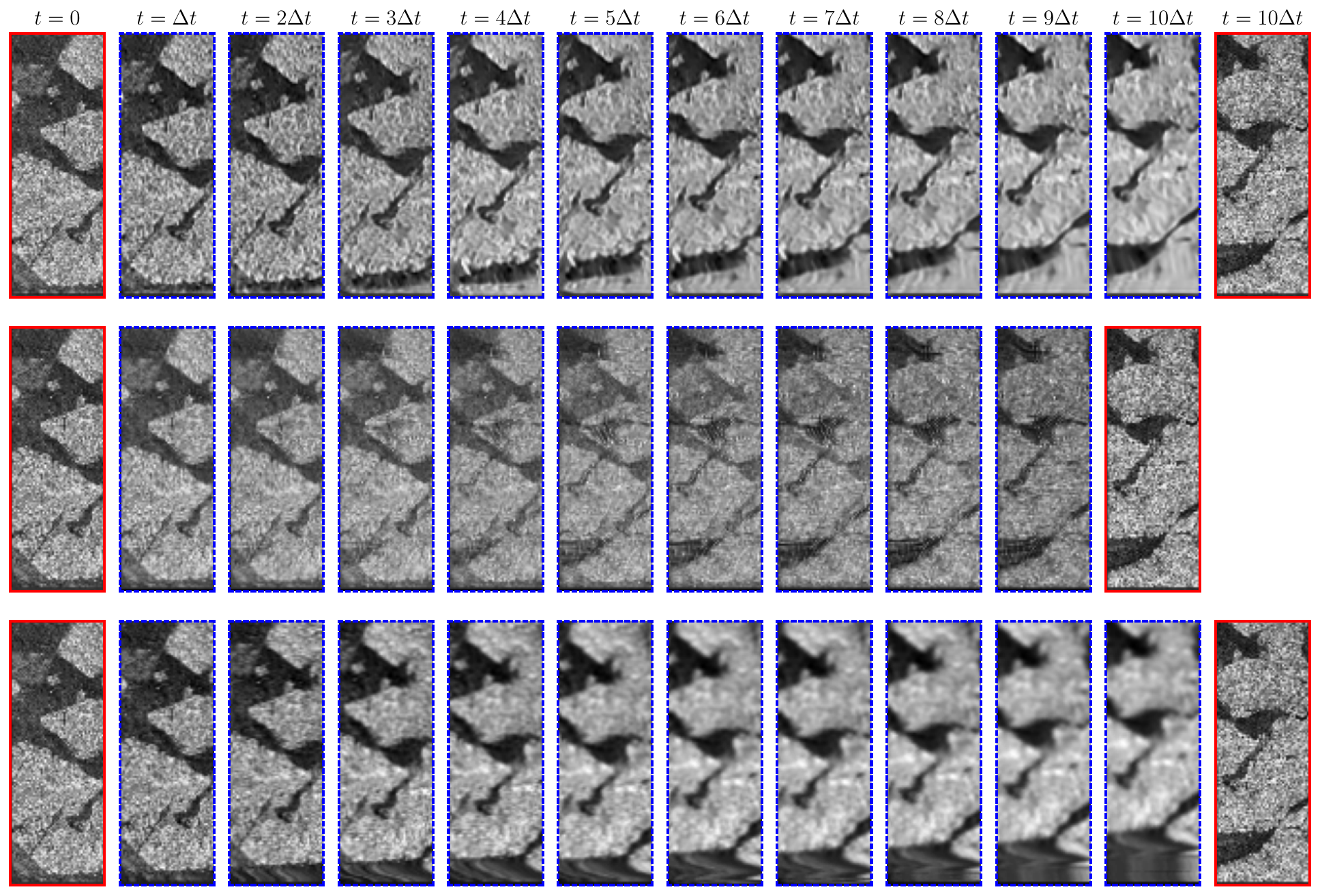}
\caption{The region near the boundary of image enclosed by white bounding box}
\end{subfigure}
\caption{The dynamics of the selected regions enclosed by the (a) yellow and (b) white bounding boxes in \Cref{fig:drift_all_methods}. Each column marks a different time $t = j\Delta t$, $j = 0,\dots,10$. (Top) our new latent space dynamics approach; (middle)  OT  \cite{zhou2022efficient}; and (bottom) direct application of PDE model in the image space.}
\label{fig:drift_boxes_all_methods}
\end{figure}

The estimated dynamics are illustrated in the first (left-most) column of \Cref{fig:drift_all_methods} (depicted as images with blue dashed borders) and the selected regions enclosed by the colored boxes are detailed in the first row of \Cref{fig:drift_boxes_all_methods}. The model effectively distinguishes shifting sea ice from stationary land.  We observe in particular that the land boundaries maintain constant values even as the ice begins to deform. The sea ice transitions diagonally across each image, as highlighted by yellow boxes, preserving intricate details like ice cracks (see \Cref{fig:drift_boxes_all_methods}(a)).  Interestingly, the sea ice smoothly recedes from the image frame, particularly toward the left boundary, as depicted in the region enclosed by each white bounding box. Additionally, novel features emerge along the lower horizontal boundary, seamlessly integrating into the image (see  \Cref{fig:drift_boxes_all_methods}(b)). This integration is a result of extrapolated latent features from the prior time step near the boundary. 

Our proposed latent space dynamics approach does not solely consider interior image features, but also accounts for external factors, thereby enabling an exploration of potential transformation scenarios leading to terminal states. As illustrated in the second (middle) column of \Cref{fig:drift_all_methods} and the second row of \Cref{fig:drift_boxes_all_methods}, this characteristic distinguishes our method from those employing optimal transport (OT), for which the goal is to reallocate pixel density within images while conserving mass over time. In this example, the OT is limited to simulate rigid transitions both within the image's interior and along its boundaries. In particular, the portion highlighted within the yellow boxes in each image demonstrates the smooth generation and disappearance of two deep cracks, as opposed to a rigid transition of the initial-state crack (see \Cref{fig:drift_boxes_all_methods}(a)). Similarly, the region near the boundary corresponding to the white boxes does not exhibit a natural transition beyond the boundary, but rather smoothly interpolates between the initial and terminal states (see \Cref{fig:drift_boxes_all_methods}(b)), which is seemingly unphysical.

We also conduct an experiment involving a direct search for advection fields on the images themselves, similar to what is done in \cite{de2019deep}. As demonstrated in the third column of \Cref{fig:drift_all_methods} and the third row of \Cref{fig:drift_boxes_all_methods}, while adapting an advection field to image dynamics captures the overall transition trend, it is not as effective in capturing fine details. 
\begin{figure}[h!]
\centering
\includegraphics[width=0.95\textwidth]{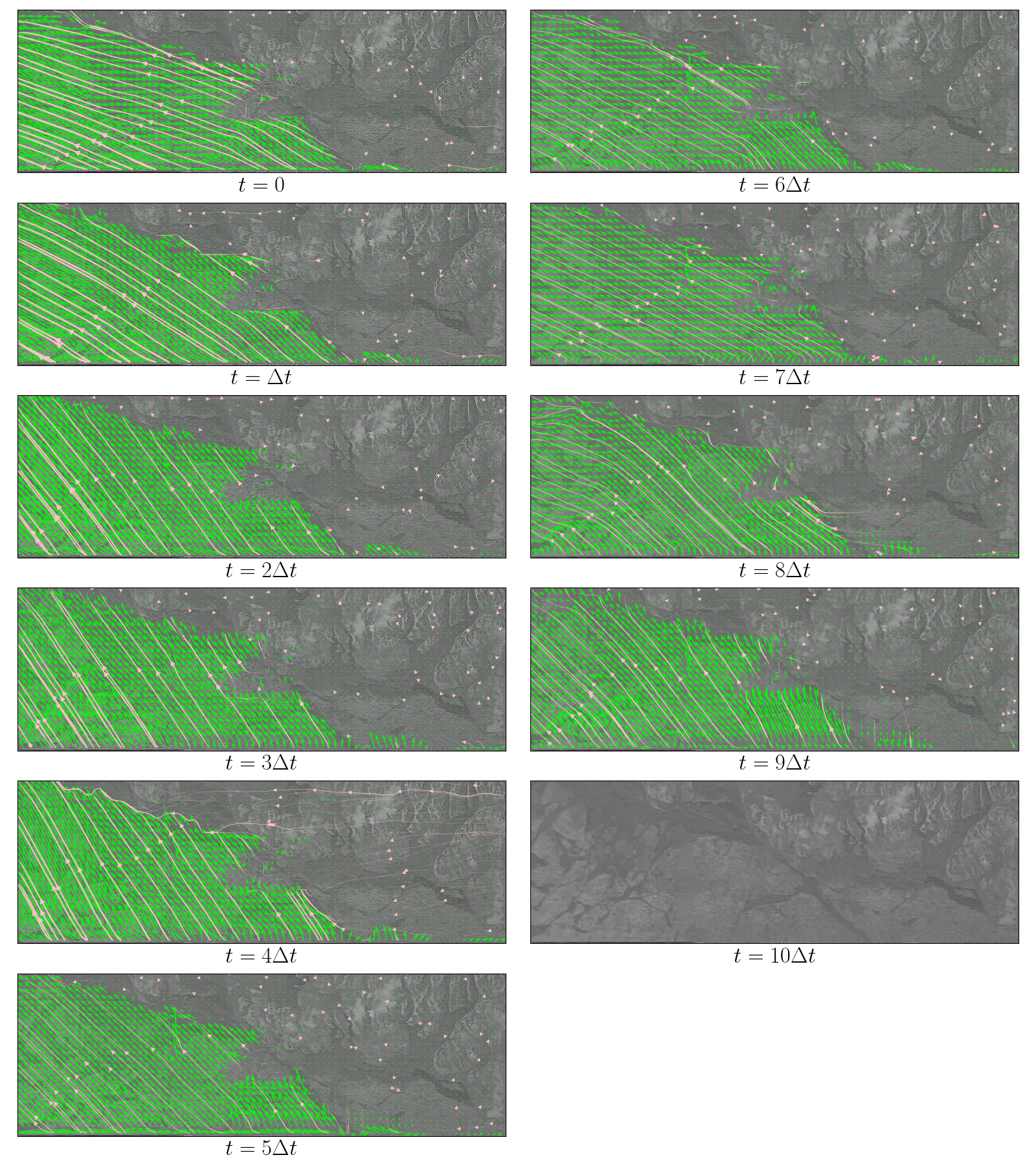}
\caption{The latent dynamics of transition-dominant image dynamics (Example 1) corresponding to the first column of \Cref{fig:drift_all_methods} along with vector fields (green) and streamlines (pink).}
\label{fig:drift_latent_sequence}
\end{figure}
Moreover, new unphysical features are generated near the boundary (see \Cref{fig:drift_boxes_all_methods}(b)). This outcome could potentially be attributed to the intricate nature of vector fields in the image space or the inadequacy of the PDE model.

We quantify image evolution by leveraging dynamic vector fields in the corresponding latent space, as depicted in \Cref{fig:drift_latent_sequence}. Spatial correlations between latent variables and images allow us to indirectly estimate image dynamics through these vector fields. The green arrows illustrate temporal feature evolution, with each denoting the direction of a feature transition for the subsequent time step. Corresponding streamlines are represented by pink lines. The vector field effectively shows the nuanced dynamics of sea ice, including its movement toward the diagonal direction and its interaction with the lower horizontal boundary and left vertical boundary (see the direction of vector fields near the boundaries, inward and outward, respectively), while static portions of sea ice and land remain unchanged. This level of detail closely aligns with the dynamics observed in the image space.

\subsection{Rotational dynamics}\label{subsec:rotation}

\begin{figure}[h!]
\centering
\includegraphics[width=0.7\textwidth]{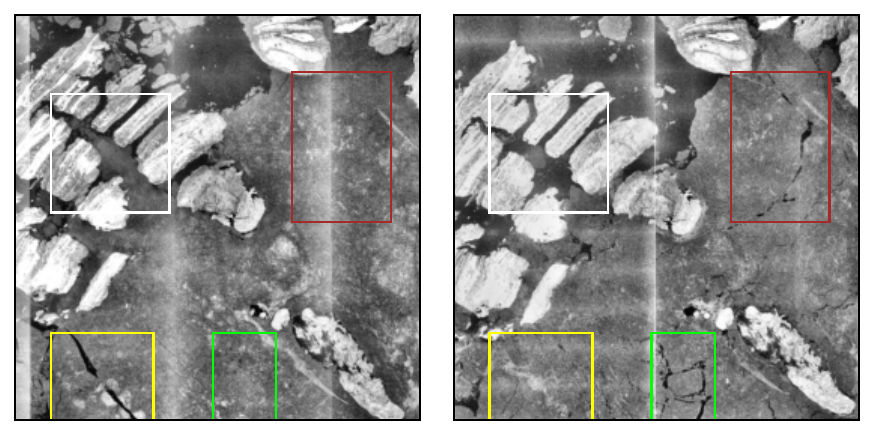}
\caption{Initial (left) and terminal (right) images for the rotational dynamics in Example 2. The colored bounding boxes indicate the regions of interest.}
\label{fig:rotation_image_initial_and_terminal}
\end{figure}

Our next scenario (Example 2) concerns rotational dynamics.  The initial and terminal images are shown in  \Cref{fig:rotation_image_initial_and_terminal}. Observe the fracture of sea ice resulting in the formation of new cracks (red and green reference bounding boxes) as well as the translation of sea ice causing cracks to merge (yellow bounding box). The data are moreover subject to measurement errors yielding low image quality, as visualized both by the vertical streaks of bright artifacts and the minor coregistration misalignments. We analyze the dynamics through eight intermediate stages with the goal to explore how sea ice movement influences the evolution and generation of cracks.

\begin{table}[h!]
\centering
\scalebox{0.95}{
\begin{tabular}{c || c | c | c} 
 & $\text{encoder}$ $\phi(\cdot;\bm{\theta}_1)$ & $\text{decoder}$ $\psi(\cdot;\bm{\theta}_2)$ & $\text{field extraction}$ $\eta(\cdot;\bm{\theta}_3)$   \\
\hline
\hline \rule{0pt}{3ex}  
$\text{hidden layers}$& $16,32,64,32,16,8$  & $\cdot$ & $\cdot$   \\
\hline
$\text{input layers}$ & $\cdot$ & $32$ &  $32$  \\
$\text{downsampling}$ & $\cdot$ & $32(2), 64(2), 128(2)$ & $32(2), 64(2), 128(2)$ \\
$\text{bottleneck}$   & $\cdot$ & $256$ & $256$ \\
$\text{upsampling}$   & $\cdot$ & $128(2), 64(2), 32(2)$ & $128(2), 64(2), 32(2)$ \\
$\text{output layers}$& $\cdot$ & $32$  & $32$  \\
\end{tabular}}
\newline
\vspace*{0.3cm}

\scalebox{0.95}{
\begin{tabular}{c | c | c | c | c | c | c} 
$N_{\text{evolution}}$ & $S_H \times S_W$ & $\lambda_{\text{AE}}$ & $\lambda_{\text{magnitude}}$ & $\lambda_{\text{smooth}}$ & $\alpha$ & $\gamma$ \\
\hline
\hline 
$8$ & $10 \times 10$ & $1.$ & $0.001$ & $0.06$ &  $0.0001$ & $0.9$ \\
\end{tabular}}

\caption{Hyperparameters for training the proposed model in Example 2. (Top) the numbers of convolutional filters in neural networks and (bottom) number of evolution steps, stride for scanning the patch over the whole image, regularization parameters, and learning parameters}
\label{table:hyperparameters for example 2}
\end{table}
Each image is comprised of $1024 \times 1024$ pixels. The images undergo a similar patch scanning as was done for the transition dynamics problem (Example 1),  with a stride of $10\times 10$, yielding $5776$ pairs of patches.  Table \ref{table:hyperparameters for example 2} provides the hyperparameter information used for training. 

\begin{figure}[H]
\centering
\includegraphics[width=1\textwidth]{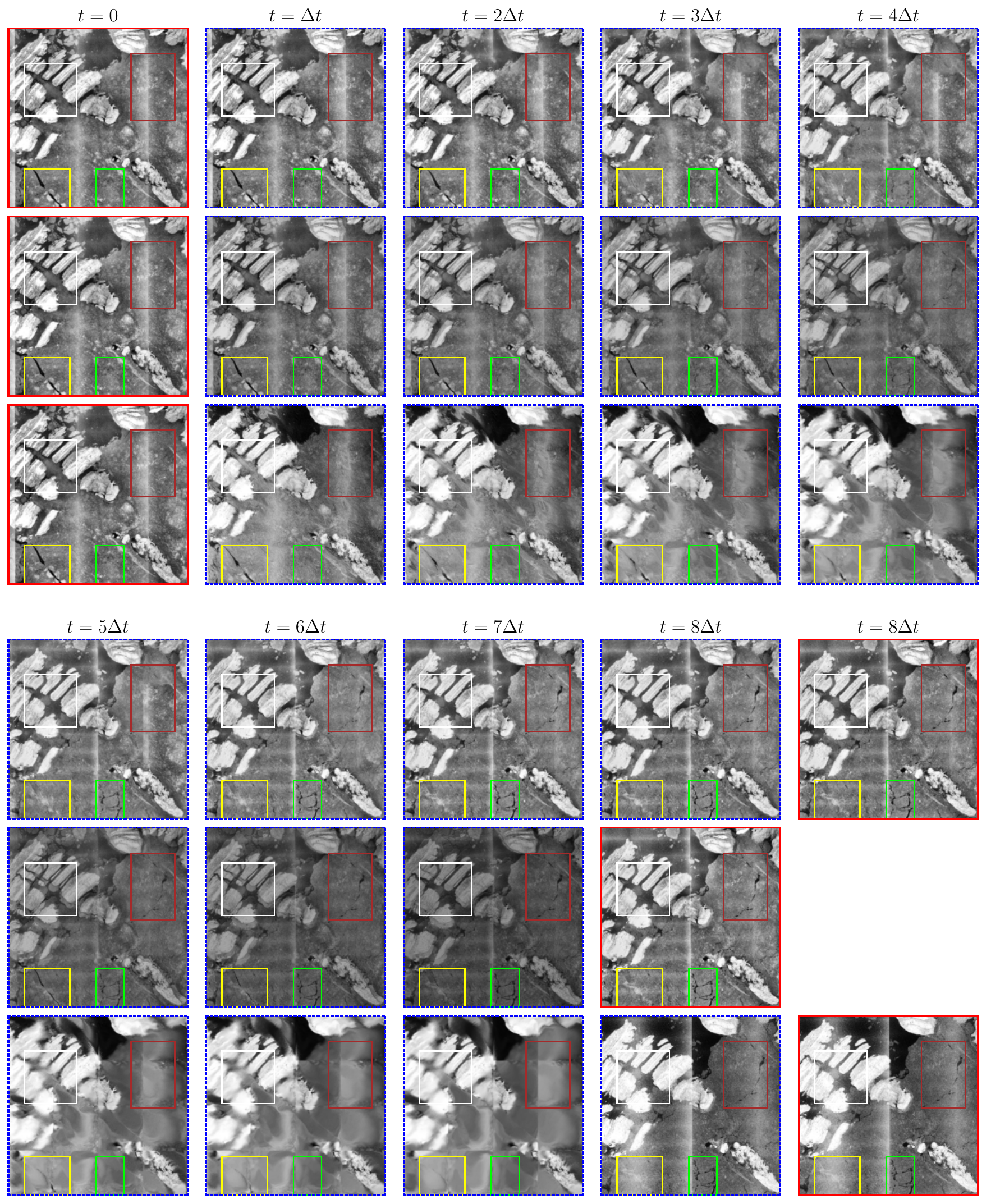}
\caption{Intermediate stages (blue dashed borders) of rotational dynamics given initial and terminal states in \Cref{fig:rotation_image_initial_and_terminal} (red borders).  (Upper) temporal sequence of images in  $[0,4\Delta t]$; (lower) temporal sequence of images in $[5\Delta t, 8\Delta t]$.  In each portion: (top) our new latent space dynamics method; (middle) OT \cite{zhou2022efficient}; and (bottom) direct application of PDE model in the image space.}
\label{fig:rotation_all_methods}
\end{figure}

\begin{figure}[H]
\begin{subfigure}[b]{1.0\textwidth}
\centering
\includegraphics[width=1\textwidth]{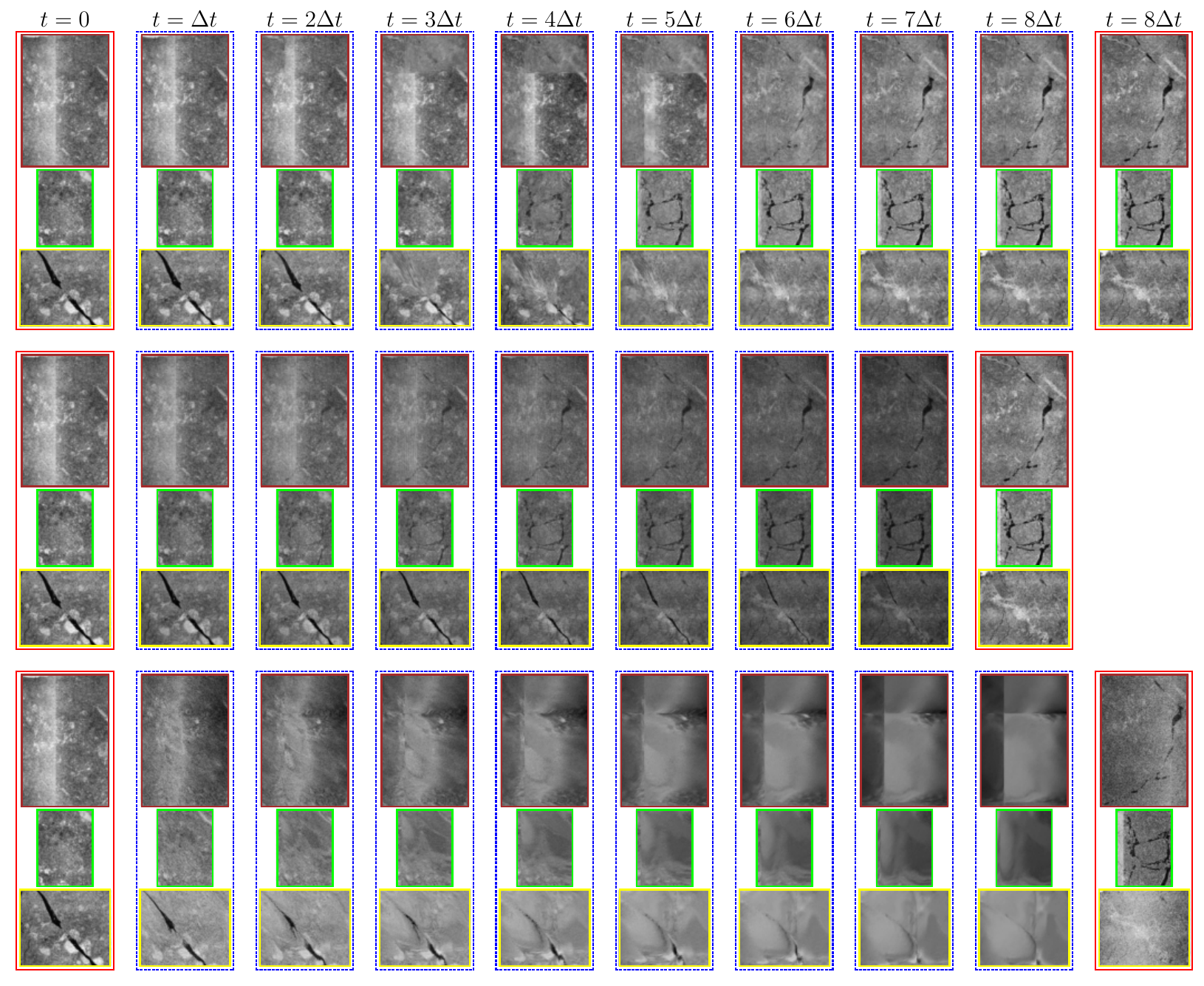}
\caption{The regions of crack generation/disappearance enclosed by yellow, green, and red bounding boxes}
\end{subfigure}
\begin{subfigure}[b]{1.0\textwidth}
\centering
\includegraphics[width=1\textwidth]{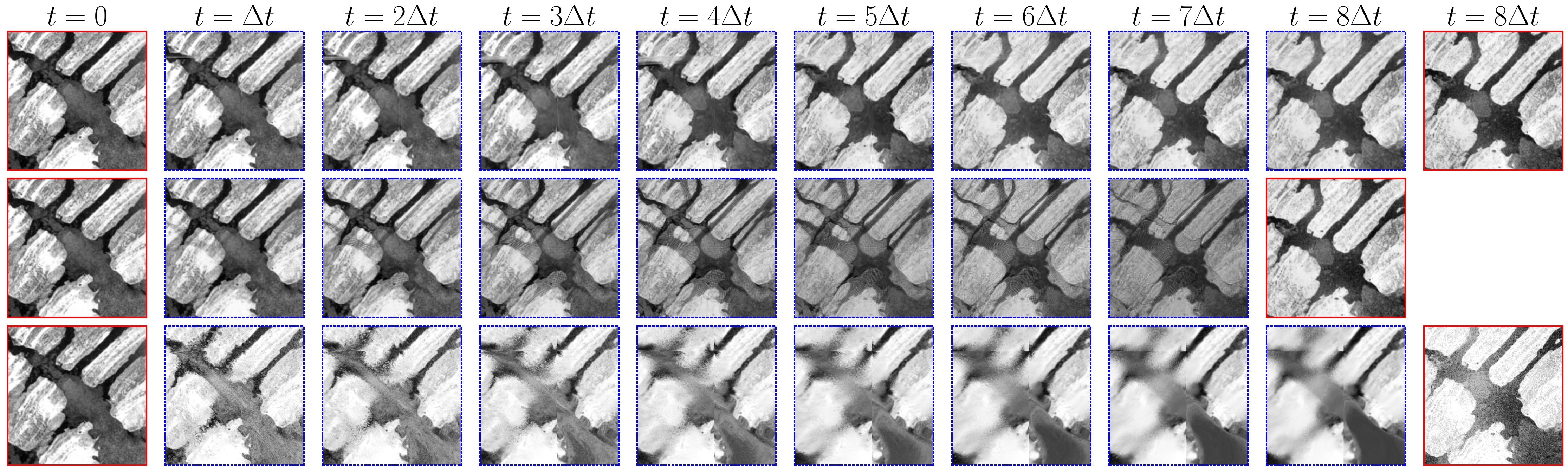}
\caption{The region enclosed by white bounding box}
\end{subfigure}
\caption{The dynamics of the (a) crack regions and (b) rotation enclosed by the colored bounding boxes in \Cref{fig:rotation_all_methods}. (top) our latent space dynamic model; (middle)  OT  \cite{zhou2022efficient}; and (bottom) direct application of PDE model in the image space.}
\label{fig:rotation_boxes_all_methods}
\end{figure}

As was done for Example 1, the model's inference is then employed in non-overlapping areas to cover the entire image, leading to the estimation of the intermediate stage dynamics.  \Cref{fig:rotation_all_methods} is split into upper ($[0,4\Delta t]$) and lower ($[5\Delta t, 8\Delta t]$) temporal sequences of images.  The first row in both sequences showcases the sequential generation and disappearance of cracks at the intermediate stages, and also demonstrates the deviation from simultaneous changes. Of note, we observe that during the initial time interval $[0,3\Delta t]$, the crack within the yellow bounding box gradually merges and vanishes. Subsequently, from $[4\Delta t, 6\Delta t]$, cracks emerge in the green  bounding boxes, followed by the appearance of a crack within the red bounding box during $[6\Delta t, 8\Delta t]$. The top row of \Cref{fig:rotation_boxes_all_methods}(a) provides additional sequential dynamics details.

These results stand in contrast to those produced using OT depicted in the second row of each sequence in \Cref{fig:rotation_all_methods}, where all cracks simultaneously and gradually evolve, either forming or disappearing over the entire time span $[0, 8\Delta t]$, which is also observed in the second row of \Cref{fig:rotation_boxes_all_methods}(a). Moreover, the dynamics resulting from OT do not demonstrate rigid movement or rotation. Instead, as can be seen in the portion enclosed in each white box spanning the time interval $[2\Delta t, 6\Delta t]$, they appear to involve local interpolation between the initial and the terminal states, as evidenced by the simultaneous presence of both states during the evolution.  This contrast is highlighted in the top and middle rows of \Cref{fig:rotation_boxes_all_methods}(b).

As also was done for Example 1, the third rows of each portion of \Cref{fig:rotation_all_methods} and \Cref{fig:rotation_boxes_all_methods} illustrate an attempt to learn the advection vector field directly within the image space. Similar to what was observed in the third column of \Cref{fig:drift_all_methods} for the transition dynamics problem, once again we see that direct application of PDEs in the image space proves ineffective in driving the image from its initial state to the terminal state.  This result further substantiates the advantage of our approach in using the latent space, as it aligns well with the physical model.

\begin{figure}[h!]
\centering
\includegraphics[width=0.84\textwidth]{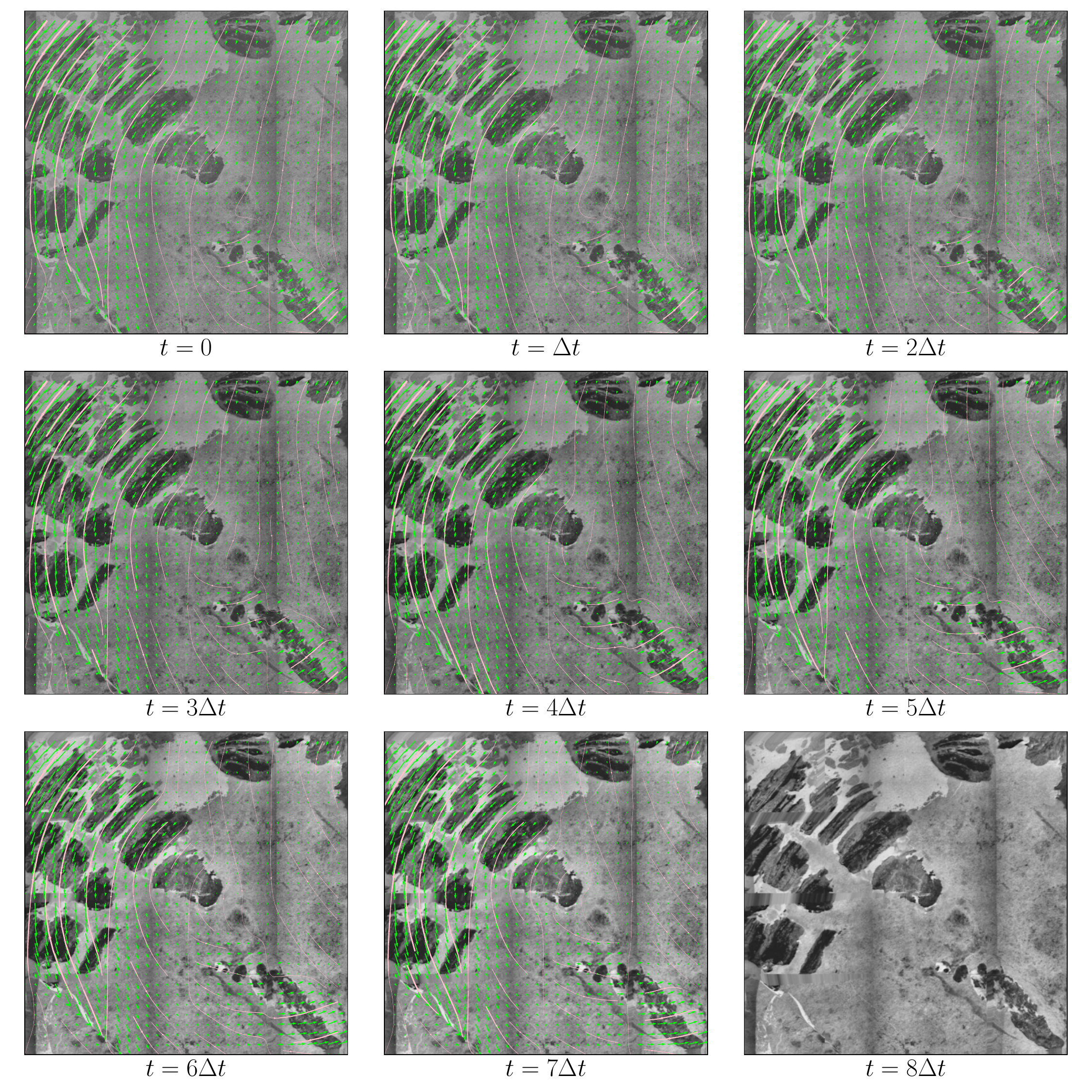}
\caption{The latent dynamics of rotational image dynamics corresponding to the first column of \Cref{fig:rotation_all_methods} along with vector fields (green) and streamlines (pink).}
\label{fig:rotation_latent_sequence}
\end{figure}

To further comprehend details of the image dynamics, we refer to the quantification of vector fields in the latent space, as depicted in \Cref{fig:rotation_latent_sequence}. By fitting an advection PDE model to the extracted latent states, the resulting field illustrates an overarching counterclockwise rotation trend throughout the evolution period.  In particular, we observe (i) A downward shift originating from the yellow bounding box compresses cracks within the box, leading to their merging. (ii) The impact then propagates through the green bounding box, fracturing the sea ice and giving rise to cracks and subsequently (iii)  a combination of upward and downward impacts around the red bounding box generates the final crack. 

Finally, we note that despite the presence of artifacts within the images, the learning of vector fields remains uninterrupted, even in proximity to latent space locations corresponding to the artifacts. We also emphasize that as demonstrated in the middle row of \Cref{fig:rotation_all_methods} (both sections), such rotational dynamics are challenging for OT methods.

\subsection{Complex dynamics with new feature generation}
\begin{figure}[H]
\centering
\includegraphics[width=0.85\textwidth]{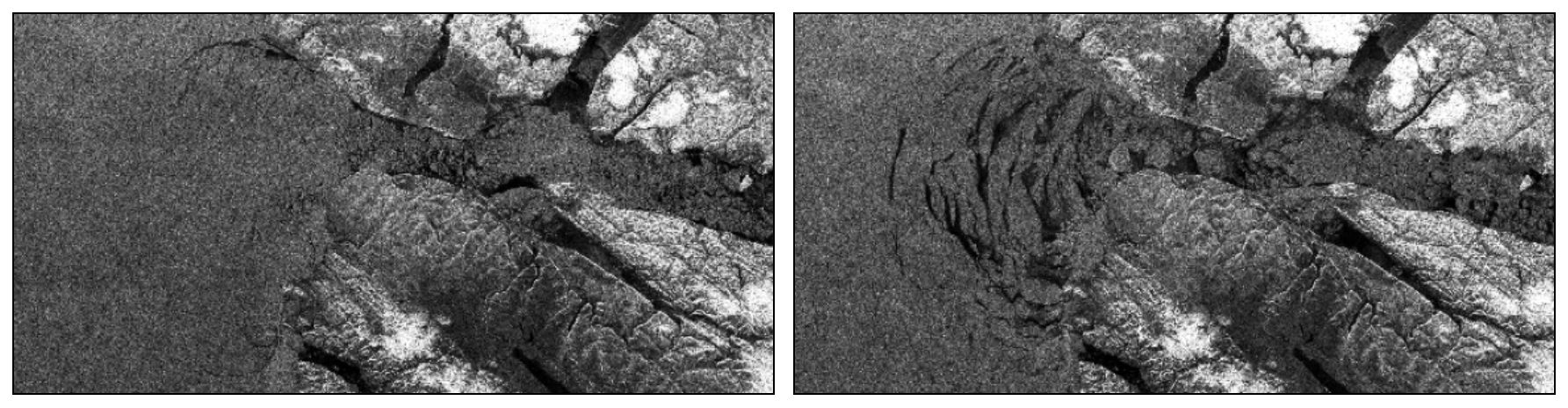}
\caption{Initial (left) and terminal (right) images for Example 3.}
\label{fig:complex_image_initial_and_terminal}
\end{figure}

\begin{table}[h!]
\centering
\scalebox{0.95}{
\begin{tabular}{c || c | c | c} 
 & $\text{encoder}$ $\phi(\cdot;\bm{\theta}_1)$ & $\text{decoder}$ $\psi(\cdot;\bm{\theta}_2)$ & $\text{field extraction}$ $\eta(\cdot;\bm{\theta}_3)$   \\
\hline
\hline \rule{0pt}{3ex}  
$\text{hidden layers}$& $16, 32, 64, 32, 16, 8$  & $\cdot$ & $\cdot$   \\
\hline
$\text{input layers}$ & $\cdot$ & $32$ &  $32$  \\
$\text{downsampling}$ & $\cdot$ & $32(2), 64(2), 128(2)$ & $32(2), 64(2), 128(2)$ \\
$\text{bottleneck}$   & $\cdot$ & $256$ & $256$ \\
$\text{upsampling}$   & $\cdot$ & $128(2), 64(2), 32(2)$ & $128(2), 64(2), 32(2)$ \\
$\text{output layers}$& $\cdot$ & $32$  & $32$  \\
\end{tabular}}
\newline
\vspace*{0.3cm}

\scalebox{0.95}{
\begin{tabular}{c | c | c | c | c | c | c} 
$N_{\text{evolution}}$ & $S_H \times S_W$ & $\lambda_{\text{AE}}$ & $\lambda_{\text{magnitude}}$ & $\lambda_{\text{smooth}}$ & $\alpha$ & $\gamma$ \\
\hline
\hline 
$10$ & $10 \times 10$ & $1.$ & $0.01$ & $0.01$ &  $0.0001$ & $0.8$ \\
\end{tabular}}
\caption{Hyperparameters for training the proposed model in Example 3. (Top) the numbers of convolutional filters in neural networks; (bottom) number of evolution steps, stride for scanning the patch over the whole image, regularization parameters, and learning parameters}
\label{table:hyperparameters for example 3}
\end{table}

\Cref{fig:complex_image_initial_and_terminal} provides a final prototype (Example 3) to test our latent space dynamics approach. 
 The figure depicts a physical structure resembling a valley in the right part of the image. The sea ice distribution initially appears to be smooth with cracks subsequently emerging, notably at the valley entrance as well as across the valley. In contrast to the previous two examples, where existing features were essentially relocated, this scenario involves the generation of entirely new features at the terminal states.  Our objective is then to understand how the model adapts latent space extraction and corresponding vector fields to generate these new features.

\begin{figure}[h!]
\centering
\includegraphics[width=1.0\textwidth]{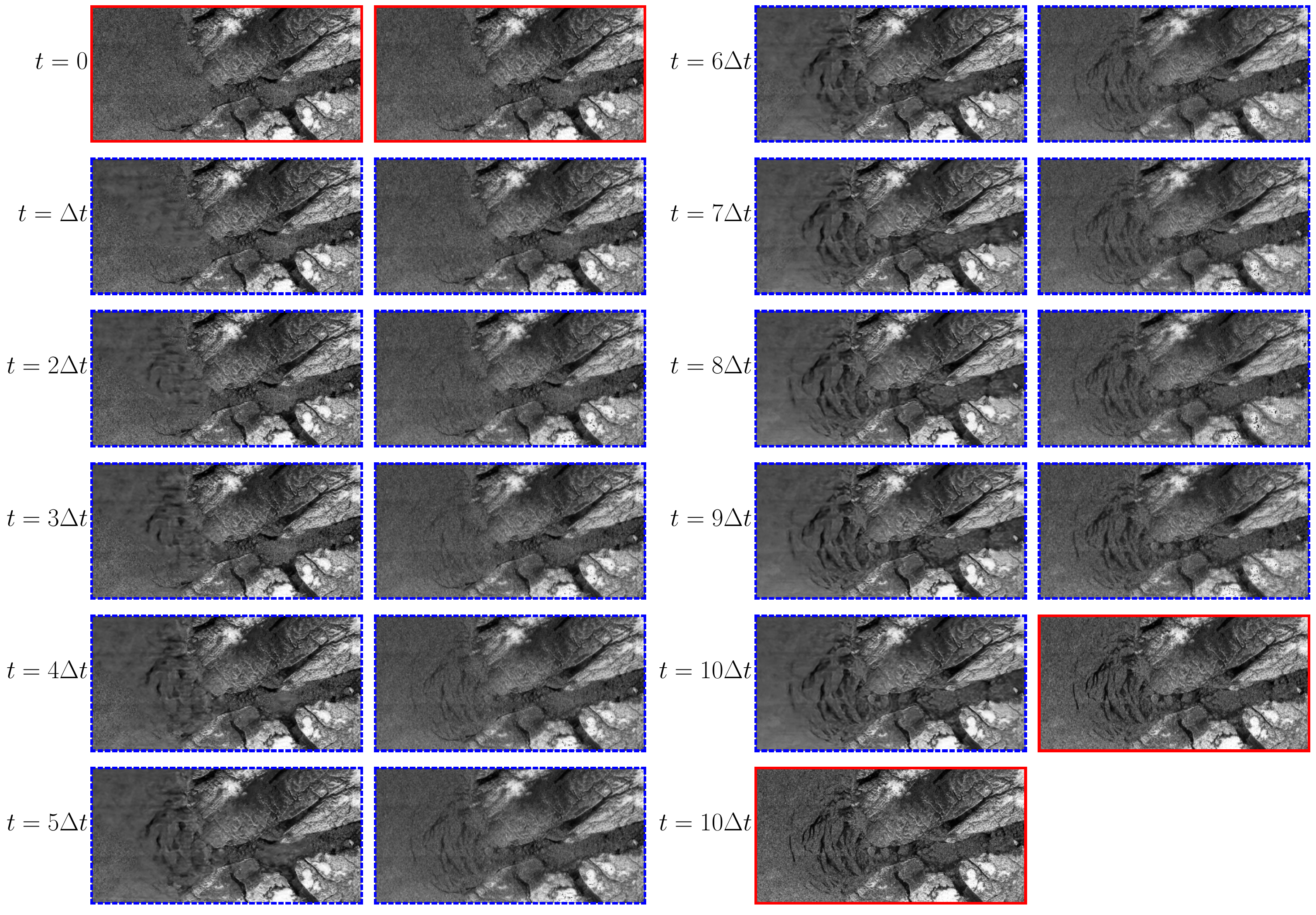}
\caption{The estimation for intermediate stages (blue dashed borders) of the dynamics given initial and terminal states in \Cref{fig:complex_image_initial_and_terminal} (red borders). (First and third  columns) our new latent space dynamics method; (second and fourth columns) OT approach \cite{zhou2022efficient}.}
\label{fig:complex_all_methods}
\end{figure}

Our model is trained with the hyperparameters detailed in Table~\ref{table:hyperparameters for example 3}, and employs pairs of patches extracted from images measuring $256 \times 512$ pixels. The first and third columns of \Cref{fig:complex_all_methods} showcase the evolution of crack generation at intermediate times $t 
 \in [\Delta t, 10\Delta t]$, where the non-overlapping inference is derived from the latent space dynamics trained model.  We observe that over this time period, the left portion in each of the images, that is,  the region outside the valley, gradually undergoes crack formation. These cracks exhibit a tendency to move rightward toward the valley. Additionally, starting at $t=4\Delta t$, cracks begin to appear and propagate into the valley.  As was done in our previous examples,  we compare these results to those obtained using OT, whch are shown in the second and fourth columns of \Cref{fig:complex_all_methods}. In this case we observe that all cracks across the image, spanning from the valley's entrance to the valley itself, gradually and concurrently become more distinct, rather than sequentially through dynamic movements. Additionally, due to disparities in total pixel density between the initial and terminal images, the normalized images exhibit variations in stationary valley regions (see bottom right corner of each frame).
 
\begin{figure}[h!]
\centering
\includegraphics[width=0.95\textwidth]{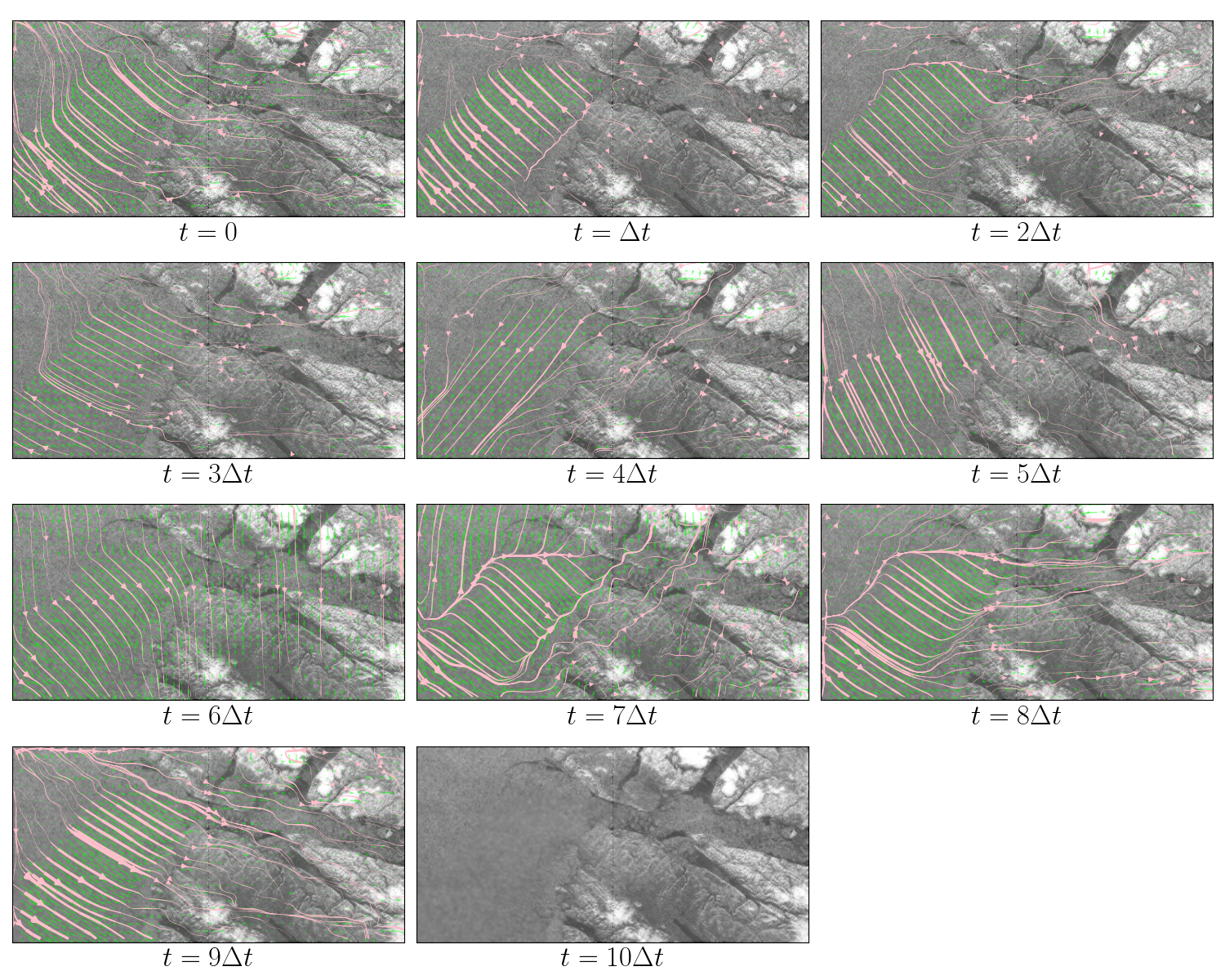}
\caption{The latent dynamics of complex image dynamics corresponding to the first column of Fig.~\ref{fig:complex_all_methods} along with vector fields (green) and streamlines (pink).}
\label{fig:complex_latent_sequence}
\end{figure}

\Cref{fig:complex_latent_sequence} provides additional insight into the evolution in latent space, offering detailed vector fields. These fields reveal a dynamic shift roughly counterclockwise over the span of $[0,6\Delta t]$. The remaining period, $[6\Delta t, 10\Delta t]$, displays discernible streamlines penetrating into the valley.

\section{Concluding remarks}\label{sec:conclusion}

This work introduces a new latent space dynamics neural network approach for estimating in-between imagery dynamics. Given the absence of temporal information in the provided data and the inherent complexity of images, addressing the unique evolution remains a significant challenge. To address this, our new method employs a PDE model to indirectly drive image evolution through a physically informed latent space. This approach is beneficial since latent dynamics can be more comprehensible when framed within PDEs. Specifically, our framework employs the advection equation to model latent evolution while maintaining spatial correlation between image and latent space. This correlation enables us to interpret image dynamics both quantitatively and qualitatively through learned advection vector fields.

We substantiate the effectiveness of our proposed method through the SAR imagery in sea ice investigation. Our method compares favorably to more commonly used OT approaches. In particular, it transcends limitations tied to total pixel conservation, allowing for interpolation beyond image boundaries to generate inflow and outflow features. Moreover, it effectively simulates new feature emergence even when the initial and terminal states exhibit discrepancies in total pixel density. Finally, our approach effectively captures rotational dynamics, a well-known challenge for OT methods \cite{villani2009optimal}.

Our work presents opportunities for approximating in-between imagery dynamics in other application domains. Specifically, we have demonstrated the power of learning patch-to-patch dynamics through continuous image scanning, implicitly accounting for global-scale features through neural networks. One possible  extension might be to directly consider different scales with various patch sizes, thereby effectively capturing the multiscale nature of image features.  This is akin to the principles underlying multigrid methods. A recent study \cite{HiPINN} shows that hierarchical training based on the different scale components of the multiscale features can expedite the training process while improving the prediction skills. Thus, it is natural to investigate whether such hierarchical learning can be applied in learning multiscale dynamics.

There are also interesting opportunities in exploring the use of other physical operators (beyond advection) within latent dynamics and in assessing their contributions to image dynamics. Different types of PDEs may exhibit varying levels of trainability. For example, evolution driven by diffusion is intertwined through neighborhood values, and is distinct from the advection case. 

Finally, incorporating any additional available information about the intermediate time periods may significantly reduce scenario uncertainty. Strategies including regularization and/or data assimilation during the encoding and decoding process could prove valuable in this context. Our proposed method provides a strong foundation for any of these future research directions.

\section*{Acknowledgments}
All authors are supported by the DoD MURI grant ONR \# N00014-20-1-2595. AG is also supported by the AFOSR grant \# FA9550-22-1-0411 and the NSF grant DMS \# 1912685.

\bibliographystyle{siamplain}
\bibliography{references}

\end{document}